%% file: argus.tex
\documentclass{article}

\PassOptionsToPackage{numbers,compress}{natbib}
\usepackage[preprint]{neurips_2026}

\usepackage[utf8]{inputenc}
\usepackage[T1]{fontenc}
\usepackage{url}
\usepackage{booktabs}
\usepackage{amsfonts}
\usepackage{amsmath}
\usepackage{amssymb}
\usepackage{nicefrac}
\usepackage{microtype}
\usepackage{xcolor}
\usepackage{graphicx}
\usepackage{algorithm}
\usepackage{algorithmic}
\usepackage{multirow}
\usepackage{subcaption}
\usepackage{enumitem}
\usepackage{tikz}
\usepackage{makecell}
\usepackage[table]{xcolor}
\usepackage{wrapfig}
\usetikzlibrary{arrows.meta, positioning, shapes.geometric}
\usepackage{tcolorbox}
\tcbuselibrary{skins, breakable, listings}
\usepackage{xcolor}
\usepackage{fancyvrb}
\definecolor{argusbg}{HTML}{F9F8F7}
\definecolor{argusborder}{HTML}{9C9594}
\definecolor{argusred}{HTML}{FF3C43}
\definecolor{argusgold}{HTML}{FFC00A}
\definecolor{argusbg}{HTML}{F9F8F7}
\definecolor{argusborder}{HTML}{9C9594}
\definecolor{arguscyan}{HTML}{507887}
\definecolor{argusred}{HTML}{FF3C43}
\usepackage{hyperref}
\hypersetup{
    colorlinks=true,
    urlcolor=blue,
}

\title{Argus: Evidence Assembly for Scalable \\ Deep Research Agents}

\author{%
  \begin{tabular}{@{}c@{}}
    \textbf{Zhen Zhang}$^{\ddagger}$\thanks{Equal contributions.}\,\,,\,
    \textbf{Liangcai Su}$^{*}$,\,
    \textbf{Zhuo Chen}$^{*}$,\,
    \textbf{Xiang Lin},\,
    \textbf{Haotian Xu},\,
    \textbf{Kaiyu Yang},\, \\[0.5pt]
    \textbf{Bo An},\,
    \textbf{Simon Shaolei Du}$^{\ddagger}$,\,
    \textbf{Lidong Bing}\thanks{Corresponding author.}\,\,,\,
    \textbf{Xinyu Wang}$^{\dagger}$\thanks{Simon Shaolei Du, Xinyu Wang and Zhen Zhang are project leaders.}
    \\[4pt]
    {\normalfont MiroMind AI}
  \end{tabular}%
}

\begin{document}

\maketitle

\input{section/00_abstract}

\input{section/01_introduction}

\input{section/03_method}

\input{section/04_training}

\input{section/05_experiments}

\input{section/02_related_work}

\input{section/07_conclusion}

\bibliographystyle{unsrtnat}
\bibliography{references}

\newpage
\appendix

\input{section/A_training}

\input{section/case_1}

\end{document}

%% file: section/00_abstract.tex
\begin{abstract}
Deep research agents have achieved remarkable progress on complex information seeking tasks. Even long ReAct style rollouts explore only a single trajectory, while recent state of the art systems scale inference time compute via parallel search and aggregation. Yet deep research answers are composed of complementary pieces of evidence, which parallel rollouts often duplicate rather than complete, yielding diminishing returns while pushing the aggregation context toward the model's limit. We propose \textbf{Argus}, an agentic system in which a Searcher and a Navigator cooperate to treat deep research as assembling a jigsaw from complementary evidence pieces, rather than brute forcing the whole answer in parallel. The Searcher collects evidence traces for a given sub-query through ReAct-style interaction. The Navigator maintains a shared evidence graph, verifying which pieces are still missing, dispatching Searchers to gather them, and reasoning over the completed graph to produce a source-traced final answer. We train the Navigator with reinforcement learning to verify, dispatch, and synthesize, while independently training the Searcher to remain a standard ReAct agent. The resulting Navigator supports rollouts with a single Searcher or many in parallel without retraining. With both Searcher and Navigator built on a 35B-A3B MoE backbone, Argus gains $5.5$ points with a single Searcher and $12.7$ points with $8$ parallel Searchers, averaged over eight benchmarks. With $64$ Searchers it reaches $86.2\%$ on BrowseComp, surpassing every proprietary agent we benchmark, while the Navigator's reasoning context stays under $21.5$K tokens.
\end{abstract}

%% file: section/01_introduction.tex
\section{Introduction}
\label{sec:intro}

Deep research agents have become a primary testbed for agentic LLM capabilities, answering complex information-seeking questions through iterative search and reasoning over web sources~\cite{openai2025deepresearch,google2025geminideepresearch,xai2025grok3,tongyi}. Even with long ReAct-style rollouts, a single trajectory explores only one sequential path through the search space, limited by what one actor can find in one pass. The current state of the art therefore scales inference-time compute in parallel: $K$ trajectories are sampled independently and then aggregated through majority voting~\cite{wang2023selfconsistency}, best-of-$N$ selection~\cite{uesato2023solving, cobbe2021training, bestofn}, or LLM-based synthesis~\cite{ParallelMuse,hu2026pacore}. Yet these gains saturate at small $K$. Deep research answers are composed of complementary pieces of evidence, which parallel rollouts often duplicate rather than complete. Each additional trajectory thus yields diminishing information gains while pushing the aggregation context toward the model's limit.

\begin{figure}[t]
  \centering
  \includegraphics[width=1.0\linewidth]{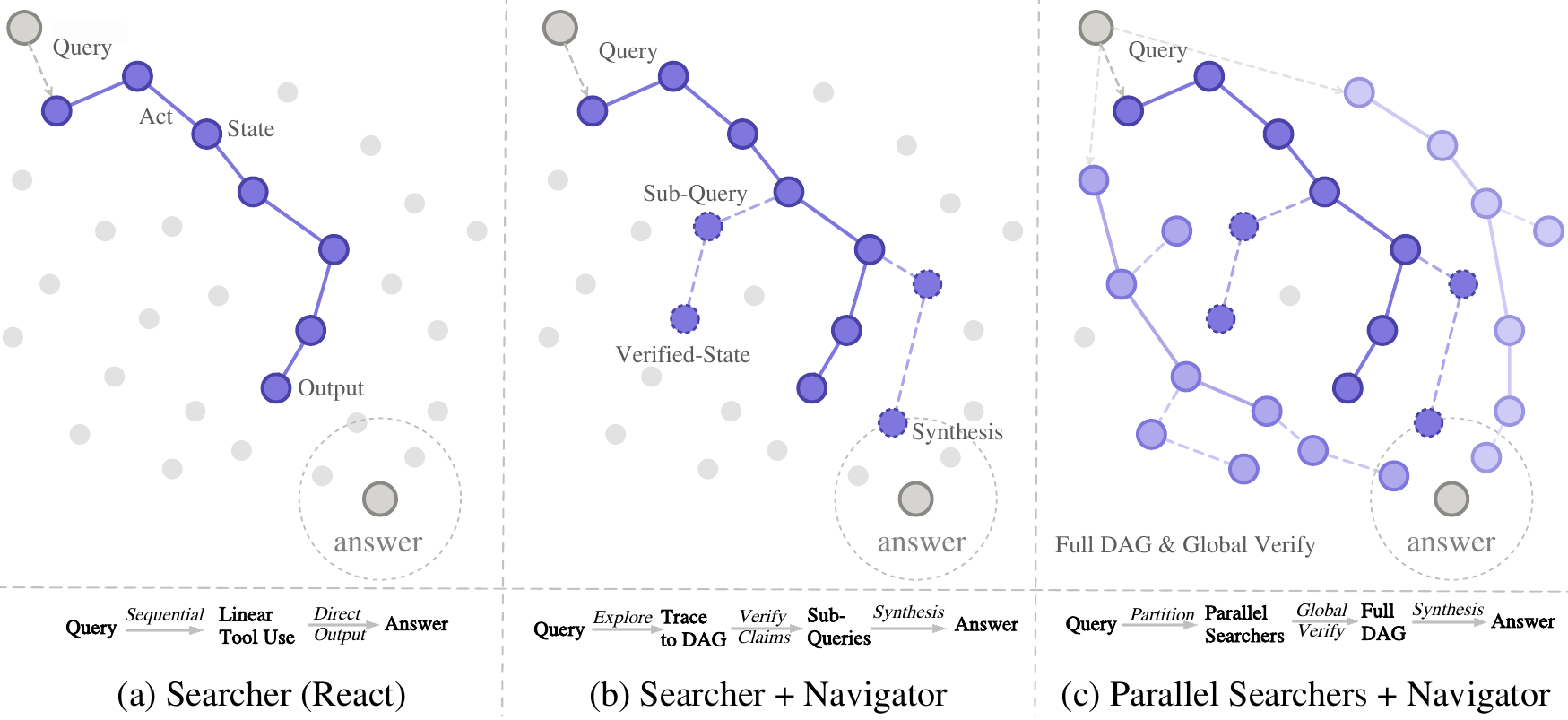}
  \caption{%
    \textbf{Argus operating modes.}
    \textbf{(a)} Standalone Searcher, single path.
    \textbf{(b)} Navigator identifies unfilled pieces and dispatches
    targeted queries.
    \textbf{(c)} Parallel Searchers each target a distinct piece.
  }
  \label{fig:overview}
\end{figure}

Fundamentally, this redundancy stems from a limitation in how the search process is represented. A ReAct-style trajectory
is a linear chain of thoughts, tool calls, and observations, produced by a single agent over one continuous rollout (Figure~\ref{fig:overview}(a)). Stacking $K$ such chains in parallel adds more chains but not more structure~\cite{brown2024large}. The result is still a
flat collection of linear traces, with no shared notion of which pieces of evidence have been gathered, which support or contradict one another, and which are still missing. Existing parallel-agent
methods inherit this flatness: self-consistency~\cite{wang2023selfconsistency},
best of-$N$~\cite{bestofn,cobbe2021training}, learned aggregation~\cite{aggagent,zeng2026pushing}, and RL-trained agent swarms~\cite{kimik25} all consume the $K$ rollouts first and select over their final answers, so gains saturate once new rollouts retrieve overlapping evidence. The compositional structure that the answer demands, with pieces that must fit together across viewpoints and sources, has no place to live in this representation.

We propose \textbf{Argus} which uses a pair of cooperating agents over a shared evidence graph to lift linear chains into a structured whole. A Searcher simply runs a single ReAct rollout and returns its trace. The Navigator orchestrates the process by incrementally building a directed acyclic graph where evidence and tentative claims become nodes while support and contradiction become edges. This graph makes missing evidence and unresolved contradictions computable. The Navigator detects these gaps and dispatches new Searchers at specific targets instead of rerunning the whole task as shown in Figure~\ref{fig:overview}(b). It continues this verify and dispatch loop until the graph is complete, seamlessly absorbing sequential or parallel trajectories as seen in Figure~\ref{fig:overview}(c). Once construction finishes, the Navigator clears its working context and reasons solely over the question and the assembled graph to synthesize a final answer. This separation keeps the reasoning context small because the graph is a compact summary so the Navigator never needs to reread raw chains. It outputs the final answer and the full graph providing a source traced reasoning path for every claim. We train the Navigator end to end with reinforcement learning ensuring the loop builds useful graphs and the reasoning step reliably extracts correct answers.

Extensive experiments demonstrate that Argus achieves SOTA accuracy on five of eight benchmarks. Built on a 35B-A3B MoE backbone for both the Searcher and the Navigator, Argus improves over the raw Searcher by $+5.5$ points on average with a single Searcher, and by $+12.7$ points with $8$ parallel Searchers. Scaling parallelism
this far exceeds the capacity of most learned aggregators, whose combiner must consume every rollout's full transcript and is capped by its own context window. Argus instead routes all $25.6$M tokens of accumulated Searcher output through the graph
and presents the Navigator with a $21.5$K token view of it, a $1{,}200{:}1$ compression that decouples Navigator context from Searcher count. Under this scaling, Argus reaches $86.2\%$ on BrowseComp at $64$ Searchers, exceeding every proprietary agent
we benchmark.

%% file: section/03_method.tex
\section{Argus: Agentic Evidence Assembly}
\label{sec:method_Argus}

\label{sec:overview}
\begin{figure}[t]
  \centering
  \includegraphics[width=1.0\linewidth]{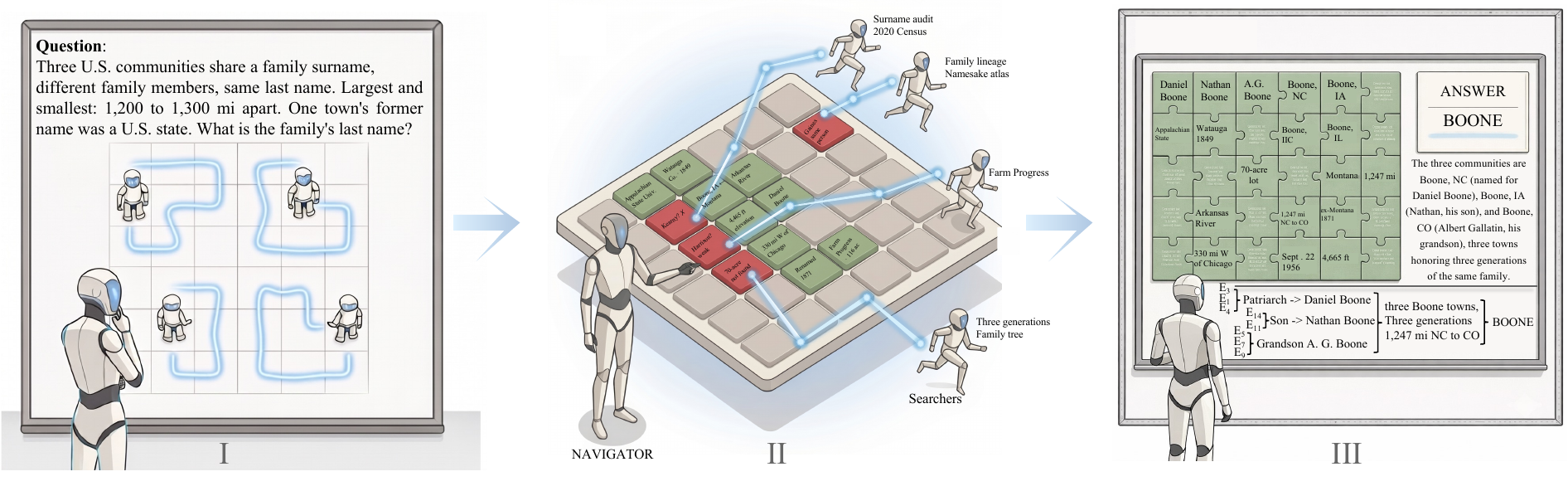}
  \caption{%
    \textbf{Argus assembles answers like a jigsaw on a
    BrowseComp-style question.}
    \textbf{(I) Parallel exploration}: Searchers execute ReAct
    rollouts.
    \textbf{(II) Navigator-guided verification}: the Navigator
    consolidates findings onto a shared evidence board
    (green: corroborated pieces; red: discarded probes) and dispatches
    Searchers at distinct gaps.
    \textbf{(III) Synthesis}: the Navigator traces each claim to its
    evidence $E_i$ and outputs the grounded final answer.
  }
  \label{fig:argus_overview}
\end{figure}

Argus consists of two cooperating agents, a \emph{Searcher} and a
\emph{Navigator}, that share a directed acyclic graph $\mathcal{G}$
of evidence and tentative claims linked by support and contradiction
edges. Given an input question $q$, Argus produces a final answer
$a$ together with the assembled graph $\mathcal{G}$ that justifies
it. Figure~\ref{fig:argus_overview} illustrates the three operating
stages.

\textbf{Step (I) Searching for evidence.}
The Navigator rewrites $q$ into one or more queries emphasizing
different angles of inquiry. In \emph{solo} mode it produces a
single rewrite; in \emph{parallel} mode several rewrites diversify
the initial coverage. The choice is a single configuration that
leaves the rest of the system unchanged. Each query is assigned to
a Searcher, which runs an independent ReAct-style rollout of
thoughts, tool calls, and observations, and returns the resulting
trace to the Navigator.

\textbf{Step (II) Verifying and assembling the graph.}
The Navigator parses each returned trace into evidence and claim
nodes and connects them into $\mathcal{G}$ with support and
contradiction edges. After each round, it inspects $\mathcal{G}$
for under-supported claims, unresolved contradictions, and aspects
of $q$ not yet addressed by any node. For each such gap, it
generates a targeted follow-up query and dispatches another
Searcher. The Navigator iterates this verify-and-dispatch loop
until $\mathcal{G}$ is sufficiently complete or the compute budget
is exhausted.

\textbf{Step (III) Synthesizing the final answer.}
Once construction terminates, the Navigator discards the working
context from the loop and reasons over $(q, \mathcal{G})$ alone to
produce the final answer $a$. Every claim involved in $a$ traces
back to evidence nodes in $\mathcal{G}$, so the output pair
$(a, \mathcal{G})$ is fully auditable.

\subsection{Searcher}

A Searcher is a stateless agent that takes a single query and returns a trajectory $H$. We adopt the ReAct framework~\cite{react}: at step $t$ the Searcher emits a thought $\tau_t$, takes an action $\alpha_t = (\alpha_t^m, \alpha_t^p)$ with $\alpha_t^m \in \{\textsc{search}, \textsc{visit}, \textsc{answer}\}$, and receives an observation $o_t$, following the action vocabulary
established in prior web-browsing agents~\cite{nakano2021webgpt, zhou2024webarena, li2025searcho1}.
\textsc{search} returns the top-10 results from a web engine, \textsc{visit} returns an extractive page summary, and \textsc{answer}
terminates the rollout with a final answer and a short rationale tying that answer to the collected evidence. The complete trajectory
$H = (\tau_0, \alpha_0, o_0, \ldots, \tau_T, \alpha_T, o_T)$ with $\alpha_T^m = \textsc{answer}$ is returned to the Navigator. A Searcher carries no state across queries, does not see $\mathcal{G}$, and does
not communicate with other Searchers, making any number of invocations independent and freely parallelizable~\cite{wang2023selfconsistency, yao2023tot}.

\subsection{Navigator}
\label{sec:navigator}

The Navigator is the agent in charge of Argus.
It maintains the shared evidence graph $\mathcal{G}$, decides what
to search for next, and produces the final answer.
We describe the three stages it runs on every problem in turn.

\paragraph{Observing trajectories and growing the graph.}
The Navigator maintains a directed acyclic graph
\begin{equation}
\mathcal{G} = (E, C, \mathcal{A}),
\quad
\mathcal{A} \subseteq (E \cup C) \times C \times \{+1, -1\},
\label{eq:graph}
\end{equation}
where $E$ is the set of \emph{evidence nodes} (raw findings retrieved by Searchers, each tagged with its source URL), $C$ is the set of \emph{claim nodes} (tentative claims a Searcher draws from one or more evidence nodes or earlier claim nodes during its rollout, including the Searcher's final answer), and each arc in $\mathcal{A}$ attaches an evidence or claim node to a claim node with a \emph{support} ($+1$) or \emph{contradict} ($-1$) label.

Unlike trees over a single agent's steps~\cite{yao2023tot} or entity graphs over
static corpora~\cite{edge2024graphrag,gutierrez2024hipporag}, $\mathcal{G}$ aggregates evidence across independent Searcher trajectories. The Navigator parses each returned $H$ into new evidence and claim nodes and attaches them via
support or contradict arcs. Evidence nodes are deduplicated at the source-URL level, preventing any single page from inflating the support count of a claim and keeping $\mathcal{G}$ a compact summary of many parallel Searchers.


After each round of returns, the Navigator labels every claim node as \emph{supported}, \emph{contradicted}, or \emph{unverified} based on its incoming arcs. The labelling is performed by the Navigator policy itself
rather than by a fixed counting rule, allowing it to weigh corroboration strength, source diversity at the URL level, and the presence of
contradicting evidence jointly. This learned criterion generalizes the multi-source corroboration principle used in atomic-fact verification~\cite{min2023factscore, wei2024longformfactuality,
chern2023factool}. The next stage targets the
unsupported claims.

\paragraph{Verifying claims and dispatching new searches.}
Once observation has settled the current state of $\mathcal{G}$,
the Navigator examines $\mathcal{G}$ as a whole and decides which
parts of it require further evidence.
The decision is not made one claim at a time.
The Navigator looks across the entire graph and produces a single
batch of verification queries
$\mathcal{V} = \{v_1, \ldots, v_m\}$, where each $v_j$ targets a
specific weakness it has identified.
This batched, graph-level verification generalizes per-claim
verification schemes such as
Chain-of-Verification~\cite{dhuliawala2024chainofverification},
Self-Refine~\cite{madaan2023selfrefine}, and
Self-Ask~\cite{press2023selfask}, in which a single agent issues
verification questions about its own outputs along a single
trajectory.
These weaknesses come in three forms.
An unverified claim prompts a query that seeks an independent
corroborating source for that claim.
A contradicted claim prompts a query that seeks authoritative
resolution of the conflict between the contradicting sources.
A region of the input question $q$ that no claim in $\mathcal{G}$
yet addresses prompts a direct query for that sub-question.
The full batch $\mathcal{V}$ is then dispatched, with one Searcher
per query, all running concurrently and writing their returned
trajectories back into $\mathcal{G}$ for the next round of
observation.
The Navigator alternates observation and verification until it emits an
end-of-loop token or the compute budget $B$ is exhausted. Termination is
a learned decision rather than a fixed threshold.

\paragraph{Synthesizing the final answer over the graph.}
Once observation and verification terminate the Navigator clears the working context accumulated during the loop and synthesizes the final answer
\begin{equation}
y^\star = \pi_{\text{syn}}(q, \mathcal{G})
\label{eq:synthesize}
\end{equation}
by reasoning over the original question $q$ together with the
completed graph $\mathcal{G}$ alone where $\pi_{\text{syn}}$ is the
Navigator synthesis policy. At this step $\mathcal{G}$ is presented
to $\pi_{\text{syn}}$ as a compact summary view rather than a raw
collection of trajectory fragments, in the spirit of graph-based
knowledge consolidation for downstream
generation~\cite{edge2024graphrag, gutierrez2024hipporag}. Evidence
is clustered by source and each claim is annotated with its
verification status and a set of derived signals such as
corroboration strength and uncertainty. This summary allows
$\pi_{\text{syn}}$ to weigh well corroborated claims more heavily
and to flag or set aside claims that remain uncertain. Because
$\mathcal{G}$ is a structured summary of every Searcher trace
integrated into it rather than a concatenation of those traces the
cost of this step grows with the size of $\mathcal{G}$ rather than
with the number or length of the underlying rollouts. Every
factual claim in $y^\star$ traces back to specific evidence nodes
and their source URLs so the pair $y^\star$ and $\mathcal{G}$ is a
fully auditable answer.

%% file: section/04_training.tex
\section{Search-Verify-Synthesize Agent Learning}
\label{sec:training}

\begin{figure}[t]
\centering
\includegraphics[width=0.85\linewidth]{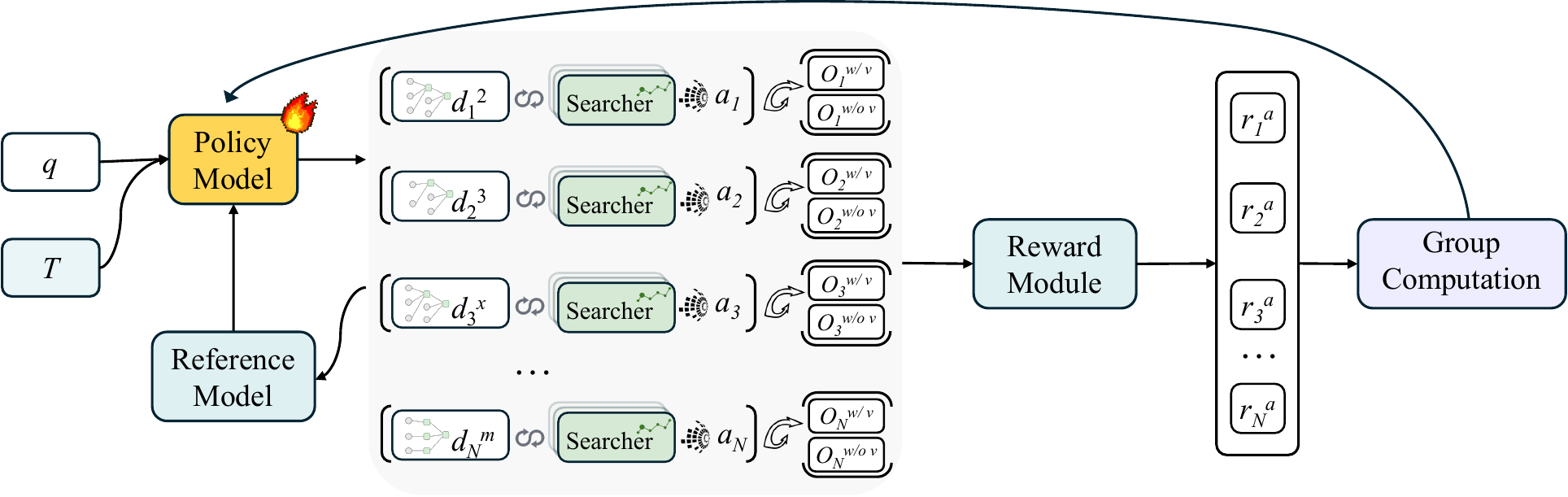}
\caption{%
\textbf{Argus GRPO training pipeline.}
Given a question $q$ and a pre-collected Searcher trajectory $T$, $\pi_\theta$ samples $N$ rollouts, each producing a full synthesis $y^\star_{\text{w/\,v}}$ over the post-verification graph and a shadow synthesis $y^\star_{\text{w/o\,v}}$ over the pre-verification graph. Their contrast yields the trajectory reward, from which GRPO computes group-relative advantages regularized by KL to a fixed reference.
}
\label{fig:training}
\end{figure}
\paragraph{Trained components.}
Argus pairs two trained components, both built on
Qwen3.5-35B-A3B~\cite{qwen35blog} (35B total, 3B active, MoE). The
Searcher is fine-tuned on SFT data produced via the
WebSailor~\cite{li2025websailor,websailorv2} pipeline. Any sufficiently capable
search agent can serve in this role, since the Navigator's
structural contribution is orthogonal to Searcher strength. The
Navigator implements the three-stage behaviour of
Section~\ref{sec:navigator} as a single policy $\pi_\theta$. It is
warm-started by SFT on the graph-construction and synthesis output
formats, then fine-tuned end to end with Group Relative Policy
Optimization (GRPO)~\cite{shao2024deepseekmathpushinglimitsmathematical} so that verification builds a graph
synthesis can convert into a correct answer. We describe the
reward, the optimization objective, and the rollout structure in
turn.


\paragraph{Reward design.}
A binary reward on final-answer correctness would credit every
trajectory that happens to land on the right answer, including
those whose verification stage contributed nothing.
We instead use a contrastive reward that isolates the causal contribution of verification, in the spirit of counterfactual credit assignment for multi-step reasoning~\cite{prm,uesato2023solving}.
For each rollout, we run synthesis twice over the same Navigator
weights.
The \emph{full synthesis}
$y^\star_{\text{w/\,v}} = \pi_{\text{syn}}(q, \mathcal{G}_{\text{post}})$
uses the post-verification graph $\mathcal{G}_{\text{post}}$, and
the \emph{shadow synthesis}
$y^\star_{\text{w/o\,v}} = \pi_{\text{syn}}(q, \mathcal{G}_{\text{pre}})$
uses the pre-verification graph $\mathcal{G}_{\text{pre}}$ before
the verification stage was run.
The shadow pass carries no gradient and is only used to compute
the reward.
Let $R_{\text{w/\,v}}$ and $R_{\text{w/o\,v}}$ be LLM-as-judge
scores of $y^\star_{\text{w/\,v}}$ and $y^\star_{\text{w/o\,v}}$
respectively.
The trajectory reward is
\begin{equation}
R_i = \mathrm{clip}\bigl(
    R_{\text{w/\,v}}
    + \lambda\,(R_{\text{w/\,v}} - R_{\text{w/o\,v}}),
    \;0,\; 1\bigr),
\qquad \lambda = 0.5.
\label{eq:contrastive-reward}
\end{equation}
The bonus term $\lambda\,(R_{\text{w/\,v}} - R_{\text{w/o\,v}})$
rewards verification queries that move the answer toward
correctness and lightly penalizes those that hurt it. We set $\lambda=0.5$ so that $R_{\text{w/ v}}$ remains the dominant
term while the contrastive bonus retains a meaningful gradient.
When the Navigator issues no verification queries, $\mathcal{G}_{\text{post}} = \mathcal{G}_{\text{pre}}$ and the
reward reduces to a clean answer-quality score.

\paragraph{GRPO objective.}
For each question the current policy $\pi_\theta$ samples a group
of $N$ rollouts $\{H_i\}_{i=1}^N$ with rewards $\{R_i\}_{i=1}^N$
computed as in Eq.~\ref{eq:contrastive-reward}.
We use the group-relative advantage
\begin{equation}
A(H_i) = R_i - \frac{1}{N}\sum_{j=1}^{N} R_j
\end{equation}
inside the PPO-clipped surrogate with a KL penalty to a fixed
reference policy $\pi_{\theta_{\mathrm{ref}}}$:
\begin{equation}
\mathcal{L}_{\mathrm{GRPO}}(\theta) =
\mathbb{E}_{H_i}\!\Bigl[
\min\!\bigl(
\rho_i\, A(H_i),\;
\mathrm{clip}(\rho_i, 1-\epsilon, 1+\epsilon)\, A(H_i)
\bigr)
\Bigr]
- \beta\, D_{\mathrm{KL}}\!\bigl(\pi_\theta \,\|\, \pi_{\theta_{\mathrm{ref}}}\bigr),
\end{equation}
where
$\rho_i = \pi_\theta(H_i)\,/\,\pi_{\theta_{\mathrm{old}}}(H_i)$ is
the importance-sampling ratio, and $\epsilon$ and $\beta$ are the
clipping threshold and KL coefficient.

\paragraph{Rollout structure.}
Each training rollout unfolds the full Argus loop on a paired question and trajectory $(q, T)$ as a single sequence. To make graph construction a learned iterative process, the observation stage builds $\mathcal{G}_{\text{pre}}$ incrementally by advancing along $T$ in a sliding window, appending evidence and claim nodes at each step. The verification stage then dispatches a batch of subsequent queries $\mathcal{V}$ to the Searcher, folding the returns into $\mathcal{G}$ to form $\mathcal{G}_{\text{post}}$. Finally, the synthesis stage generates $y^\star_{\text{w/\,v}}$ and $y^\star_{\text{w/o\,v}}$. Gradients are solely applied to tokens generated by the Navigator. The trajectory $T$, verification returns, and other external inputs are masked. 

Crucially, while training relies on a single Searcher trajectory $T$ per question (sampling $N$ Navigator rollouts for GRPO), the Navigator operates strictly on $q$ and the state of $\mathcal{G}$. This abstraction makes the policy invariant to the initial Searcher count. Consequently, a policy trained on single trajectories transfers directly to inference configurations with parallel Searcher swarms, which we verify empirically in Section~\ref{sec:scaling_exp}.

%% file: section/05_experiments.tex
\section{Experiments}
\label{sec:exp}
\subsection{Experimental Setups}
\label{sec:setup}

\paragraph{Benchmarks.}
We evaluate Argus on eight benchmarks spanning the difficulty range
relevant to deep research agents.
\textbf{BrowseComp}~\cite{browsecomp} and its Chinese counterpart
\textbf{BrowseComp-ZH}~\cite{browsecompzh} probe multi-step web browsing on adversarially constructed factual questions that resist
single-hop search. \textbf{xbench DeepSearch-2510}~\cite{chen2025xbench} targets deep search and tool use through professionally annotated Chinese tasks with dynamic updates. \textbf{GAIA}~\cite{gaia} stresses general assistant capabilities that combine tool use, multi-hop reasoning, and web search across
real-world question types. \textbf{SEAL-0}~\cite{sealqa} is the main challenge track of SealQA, designed to defeat search-augmented reasoning that relies on a single retrieval step.
\textbf{Humanity's Last Exam}~\cite{hle} probes the frontier of expert-level knowledge across science, mathematics, law, and
medicine. \textbf{FrontierScience-Olympiad}~\cite{frontierscience} targets
Olympiad-difficulty problems in physics, chemistry, and biology, written and verified by competition-level experts, while
\textbf{FrontierScience Research}~\cite{frontierscience} extends
this to open-ended PhD-level research sub-problems, probing scientific reasoning under ambiguity rather than fixed competition constraints. Together these eight benchmarks cover short-form factual lookup, multi-hop synthesis, and expert-level reasoning, the breadth Argus is designed to handle.

\paragraph{Compared systems.}
We compare Argus against three baseline groups evaluated on the same benchmark suite using metrics detailed in Appendix~\ref{app:evaluation}. The first group is the proprietary
frontier, comprising GPT-5.2~\cite{openai2026gpt52}, Claude-4.6-Opus~\cite{anthropic2026claude46},
Gemini-3.1-Pro~\cite{google2025geminideepresearch}, and Seed-2.0-Pro~\cite{seed2026modelcard}. The second is a panel of strong open-source agents, including GLM-5.0~\cite{zeng2026glm}, Kimi-K2.6~\cite{kimi_k26}, Qwen3.5-35B-A3B~\cite{qwen35blog}, Qwen3.5-397B-A17B~\cite{qwen35blog}, and DeepSeek-V4-Pro-Max~\cite{deepseekv42026}. The third is the prior open-source deep research agents that target the same task family, Tongyi-DeepResearch~\cite{tongyi} and MiroThinker-1.7~\cite{mirothinker}. Numbers for these baselines are taken from their respective official reports where available, with entries marked $\dagger$ in Table~\ref{tab:main} reproduced by us using only \texttt{search} and \texttt{visit} actions. All Argus numbers are means over three runs with different seeds. For clarity, we omit per-cell standard deviations, which remain consistently low ($\le 0.73\%$) across three independent Argus runs. We report Argus in two configurations sharing a single Navigator and a single Searcher base, both built on Qwen3.5-35B-A3B with the Searcher fine-tuned via the WebSailor-v2~\cite{websailorv2} pipeline. Searcher runs the fine-tuned Searcher alone as a plain ReAct agent without the Navigator. Argus (Solo) adds the Navigator's verify-and-dispatch loop on top of a single initial Searcher. Argus (Parallel) dispatches $K{=}8$ initial Searchers per question, with the Navigator orchestrating verification across shared graph. 

\subsection{Main Results}
\label{sec:main_results}

\begin{table}[t]
\scriptsize
\centering
\caption{%
Main results on eight complex information-seeking benchmarks. $\dagger$~Reproduced by us using only \texttt{search} and \texttt{visit} actions, without context management. $\ddagger$~Original paper evaluates on a 100 question subset; Argus on the full set. Other numbers are from official reports.
}
\label{tab:main}
\setlength{\tabcolsep}{4pt}
\renewcommand{\arraystretch}{1.2}
\resizebox{\textwidth}{!}{%
\begin{tabular}{l|cccccccc}
\toprule
\textbf{Backbone}
  & \makecell[c]{\textbf{Browse}\\\textbf{Comp}}
  & \makecell[c]{\textbf{Browse}\\\textbf{Comp-ZH}}
  & \textbf{GAIA}
  & \textbf{Seal-0}
  & \makecell[c]{\textbf{x-bench}\\\textbf{DeepSearch}\\\textbf{-2510}}
  & \makecell[c]{\textbf{Humanity's}\\\textbf{Last Exam}}
  & \makecell[c]{\textbf{FrontierScience}\\\textbf{Olympiad}}
  & \makecell[c]{\textbf{FrontierScience}\\\textbf{Research}} \\
\midrule
\rowcolor{gray!15}\multicolumn{9}{c}{\emph{\textbf{Proprietary Agents}}} \\
\midrule
GPT-5.2              & 65.8 & ---  & ---  & ---  & --- & 45.5  & 77.1  & 25.2  \\
Claude-4.6-Opus      & 83.7 &  66.8$^\dagger$ & 75.0$^\dagger$  & 50.0$^\dagger$  & ---& \textbf{53.1} & 73.0$^\dagger$  & 23.3$^\dagger$  \\
Gemini-3.1-Pro       & \textbf{85.9} & 74.0$^\dagger$  & 80.6$^\dagger$  & 42.5$^\dagger$  & 53.0& 51.4 & 76.7$^\dagger$   & 20.0$^\dagger$  \\
Seed-2.0-Pro         & 77.3 & 82.4 & 79.6$^\dagger$  & 49.5 & ---& 54.2 & 74.0 & \textbf{33.4}$^\dagger$  \\
\midrule
\rowcolor{gray!15}\multicolumn{9}{c}{\emph{\textbf{Open-Source Agents}}} \\
\midrule
GLM-5.0            & 75.9  & 73.0  & 70.0$^\dagger$  & 33.3$^\dagger$  & ---& 50.2  & 62.0$^\dagger$  & 8.3$^\dagger$  \\
Kimi-K2.6            & 83.2  & ---  & 78.6$^\dagger$  & 42.0$^\dagger$  & ---& 54.0  & ---  & ---  \\
Qwen3.5-35B-A3B      & 42.1$^\dagger$ & 47.8$^\dagger$  & 80.0$^\dagger$ & 43.2$^\dagger$ & ---& 39.5$^\dagger$ & 68.0$^\dagger$ & 3.3$^\dagger$  \\
Qwen3.5-397B-A17B    & 78.6 & 70.3 & ---  & 46.9 & ---& 48.3 & 60.6 & 11.7$^\dagger$    \\
DeepSeek-V4-Pro-Max  & 83.4 & --- & 65.3$^\dagger$  & --- & ---& 48.2 & 75.0$^\dagger$ & ---  \\
\midrule
\rowcolor{gray!15}\multicolumn{9}{c}{\emph{\textbf{Open-Source Deep Research Agents}}} \\
\midrule
Tongyi-DeepResearch  & 43.4  & 46.7  & 70.9  & ---  & 55.0 & 32.9  & ---  & ---  \\
MiroThinker-1.7      & 74.0 & 75.3 & 82.7 & 53.0 & 62.0& 42.9 & 71.5 & 8.8$^\dagger$  \\
\midrule
\rowcolor{gray!15}\multicolumn{9}{c}{\emph{\textbf{Parallel Agents}}} \\
\midrule
Tongyi-DeepResearch Heavy~\cite{zeng2026pushing}
  & 69.0$^\ddagger$ & 55.0 & 72.8 & ---
  & ---& --- & --- & ---  \\
GLM-4.5 Heavy~\cite{zeng2026pushing}
  & 54.0$^\ddagger$ & 49.0 & 66.0 & ---
  & ---& --- & --- & ---  \\
\textbf{Searcher-35B-A3B}
  & 55.0 & 62.3 & 84.5 & 48.9
  & 62.0& 43.2 & 72.0 & 5.4  \\
\textbf{Argus-35B-A3B (Solo)}
  & 62.2 & 74.4 & 88.0 & 53.2
  & 67.0& 44.2 & 75.0 & 13.2  \\
\textbf{Argus-35B-A3B (Parallel)}
  & 74.5 & \textbf{83.4} & \textbf{93.2} & \textbf{56.2}
  & \textbf{73.0}& 49.8 & \textbf{80.0} & 25.0  \\
\bottomrule
\end{tabular}}
\end{table}

Table~\ref{tab:main} compares Argus against three reference groups across eight benchmarks spanning English and Chinese deep search (BrowseComp, BrowseComp-ZH, and xbench-DeepSearch), tool-use multistep reasoning (GAIA, Seal-0), and frontier scientific problem solving (HLE, FrontierScience Olympiad and Research). Argus-Parallel leads on five of eight benchmarks, posting state-of-the-art results on BrowseComp-ZH (83.4), GAIA (93.2), Seal-0 (56.2), xbench-DeepSearch-2510 (73.0), and FrontierScience Olympiad (80.0), with the GAIA and Seal-0 margins exceeding the strongest proprietary agent by 12.6 and 6.2 points. On the remaining columns it stays within close reach of the proprietary frontier under a bounded inference budget, and pushing parallelism to 64 initial Searchers raises BrowseComp accuracy to 86.2\% (see Section~\ref{sec:scaling_exp}), exceeding every proprietary agent we benchmark. The pattern holds across question style, language, and problem domain, indicating that the compositional Navigator generalizes from open-ended browsing to structured technical questions without benchmark-specific tuning.

Two finer breakdowns deserve emphasis. Argus-Solo, which uses a single initial Searcher together with the verify-and-dispatch loop, already outperforms every open-source baseline on five of eight benchmarks and exceeds the strongest proprietary agent on GAIA ($88.0$), Seal-0 ($53.2$), and xbench-DeepSearch (67.0), which shows that most of the headline gain comes from compositional verification rather than parallel sample averaging. Argus-Parallel then extends this lead on every column, adding an average of 7.2 points over Argus-Solo with the largest gains on BrowseComp (+12.3) and FrontierScience Research (+11.8), demonstrating that compositional verification and parallel evidence gathering combine constructively rather than producing diminishing returns.


\subsection{Analysis}
\label{sec:scaling_exp}

\begin{wrapfigure}[17]{r}{0.5\textwidth}
  \centering
  \includegraphics[width=0.45\textwidth]{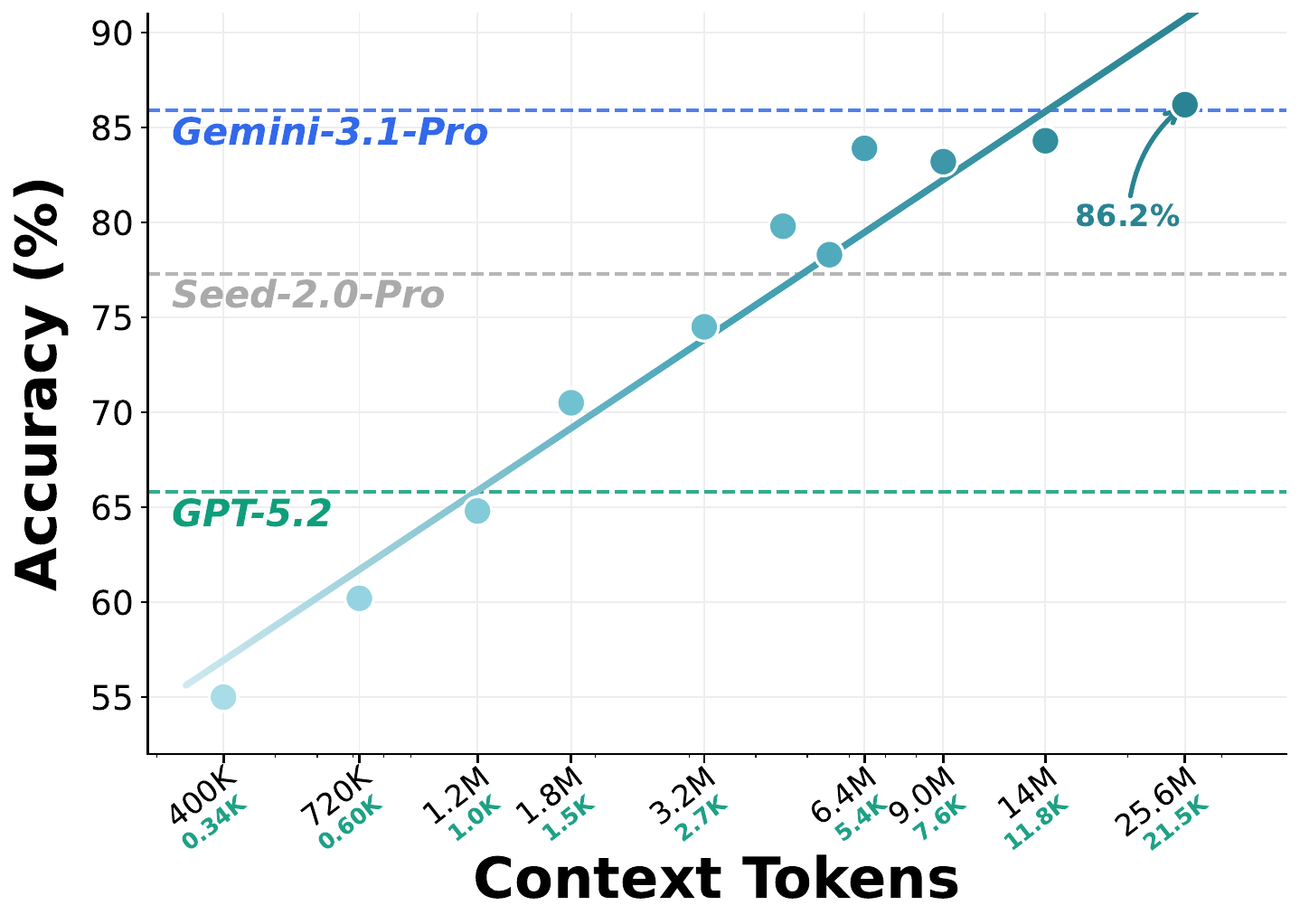}
  \caption{%
    Accuracy on BrowseComp scales log-linearly with aggregation context budget, surpassing Gemini-3.1-Pro at 64× base compute.}
  \label{fig:scaling_curve}
\end{wrapfigure}

\textbf{Scaling Behavior under Increased Budgets.}
Figure~\ref{fig:scaling_curve} plots BrowseComp accuracy against the Searcher's cumulative token consumption as we sweep parallelism and per-Searcher budget. Across the eleven configurations spanning two orders of magnitude of Searcher tokens, accuracy climbs monotonically from $55.0\%$ at $0.4$M tokens to $86.2\%$ at $25.6$M tokens, while the Navigator's synthesis context grows only from $0.34$k to $21.5$k tokens, with no sign of flattening at the rightmost point. A logarithmic fit captures the trend cleanly, suggesting further compute would still yield meaningful gains rather than hitting a hard ceiling. This follows from Argus's compositional design, where additional rollouts surface fresh evidence for the Navigator to assemble rather than duplicate guesses to be aggregated.

This decoupling is crucial. Most agentic systems hit a context wall long before exhausting compute limits. Argus instead restricts the bottleneck to the Searcher. The 21.5k token graph view at the largest budget compresses accumulated Searcher output by roughly 1200 to 1. This comfortably fits the 128k context limit of the Navigator. Parallelism thus translates directly into accuracy without inflating reasoning input. At its largest configuration Argus reaches 86.2\% on BrowseComp. It surpasses strong proprietary agents under a bounded inference budget despite using an open Searcher and a single Navigator.

\textbf{Generalization across Searcher Backbones}.
\label{sec:searcher_generalization} Table~\ref{tab:searcher_generalization} pairs the same Navigator with three different Searcher backbones, spanning one open-weight model and two proprietary systems, and compares four inference
configurations on BrowseComp. Argus-Parallel attains the highest accuracy on every backbone, with margins of $+12.3$, $+9.5$, and $+3.8$ over the next-best configuration on the three backbones respectively. The Navigator was trained with Searcher-35B-A3B in the loop, yet when dropped onto DeepSeek and Seed-2.0-Pro without any retraining it still produces a positive lift on every backbone, demonstrating zero-shot transfer of the verify-and-dispatch behaviour across heterogeneous Searcher distributions.

\begin{table}[!h]
\footnotesize
\centering
\caption{%
  Argus with different Searcher backbones on BrowseComp
  (Avg Pass@1 \%).
}
\label{tab:searcher_generalization}
\begin{tabular}{lccccccc}
\toprule
\textbf{Searcher backbone}
  & \textbf{Searcher}
  & \textbf{Argus-Solo}
  & \textbf{\makecell{Majority-Vote \\ ($K{=}8$)}}
  & \textbf{\makecell{LLM-Aggregation \\ (35B-$K{=}8$)}}
  & \textbf{\makecell{Argus-Parallel \\ ($K{=}8$)}} \\
\midrule
Searcher-35B-A3B & 55.0& 62.2 & 56.2  & 56.5  & \textbf{74.5}  \\
DeepSeek-V4-Flash-Max   & 64.0 & 68.0  & 60.0 & 69.0 & \textbf{78.5}  \\
Seed-2.0-Pro    & 70.2 & 78.6  & 67.0 & 73.8 & \textbf{82.4}  \\
\bottomrule
\end{tabular}
\end{table}

Argus-Solo, which uses a single initial Searcher together with the verify-and-dispatch loop, exceeds Majority-Vote at $K{=}8$ on every backbone by $+6.0$, $+8.0$, and $+11.6$ points respectively, showing that structured synthesis through verification dominates simple answer-level aggregation even when the latter is given eight independent rollouts. The same pattern holds against the stronger LLM-Aggregation baseline, which uses a $35$B model to consolidate the eight rollouts. Argus-Parallel exceeds LLM-Aggregation on every backbone by $+18.0$, $+9.5$, and $+8.6$ points, indicating that the structured graph view consumed by the Navigator extracts substantially more from the same eight rollouts than a free-form aggregation prompt does.

\paragraph{Ablation on graph representation.}
\begin{wraptable}{r}{0.57\textwidth}
\vspace{-1.0em}
\centering
\footnotesize
\setlength{\tabcolsep}{4pt}
\renewcommand{\arraystretch}{1.45}
\begin{tabular}{l p{0.46\linewidth} c}
\toprule
\textbf{Variant} & \textbf{What the Navigator sees} & \textbf{BrowseComp} \\
\midrule
Full DAG &
{\scriptsize $\mathcal{G}=(E, C, \mathcal{A})$ with
$\mathcal{A}\!\subseteq\!(E\!\cup\!C)\!\times\!C\!\times\!\{\!+\!1,\!-\!1\!\}$;
\texttt{status}$\in\!\{$sup, con, unv$\}$;
corroboration strength}
& \textbf{74.5} \\
\midrule
Bare graph &
{\scriptsize $\mathcal{G}=(E, C, \mathcal{A}')$ with
$\mathcal{A}'\!\subseteq\!(E\!\cup\!C)\!\times\!C$ \emph{(no $\pm 1$ label)};
coarse \texttt{status} only}
& 72.0 \\
\midrule
Text only &
{\scriptsize $\text{concat}(E_1, C_1, E_2, C_2, \ldots)$ in extraction order;
no $\mathcal{A}$, no \texttt{status}}
& 69.3 \\
\bottomrule
\end{tabular}
\caption{Graph representation ablation ($K{=}8$, BrowseComp).
All variants share identical Searcher rollouts; only the input the Navigator's synthesis stage differs.}
\label{tab:graph-ablation}
\vspace{-0.5em}
\end{wraptable}
To isolate the contribution of the structured evidence graph,
we ablate the input the Navigator's synthesis stage under
Argus-Parallel ($K{=}8$). \emph{Full DAG} is the schema in Eq.~\ref{eq:graph}, with typed support/contradict edges and per-claim verification status. \emph{Bare graph} retains the node sets $E$, $C$ and the edge connectivity in $\mathcal{A}$, but strips the $\{+1, -1\}$ labels and all node-level annotations. \emph{Text only} discards $\mathcal{G}$, joining evidence and claim text in extraction order. Performance increases monotonically: graph topology alone (Bare graph vs Text only) accounts for $+2.7$ points, and adding typed edges and annotations (Full DAG vs Bare graph) 
contributes $+2.5$, totaling $5.2$ points from the structured representation.

\begin{figure}[t]
  \centering
  \includegraphics[width=0.95\linewidth]{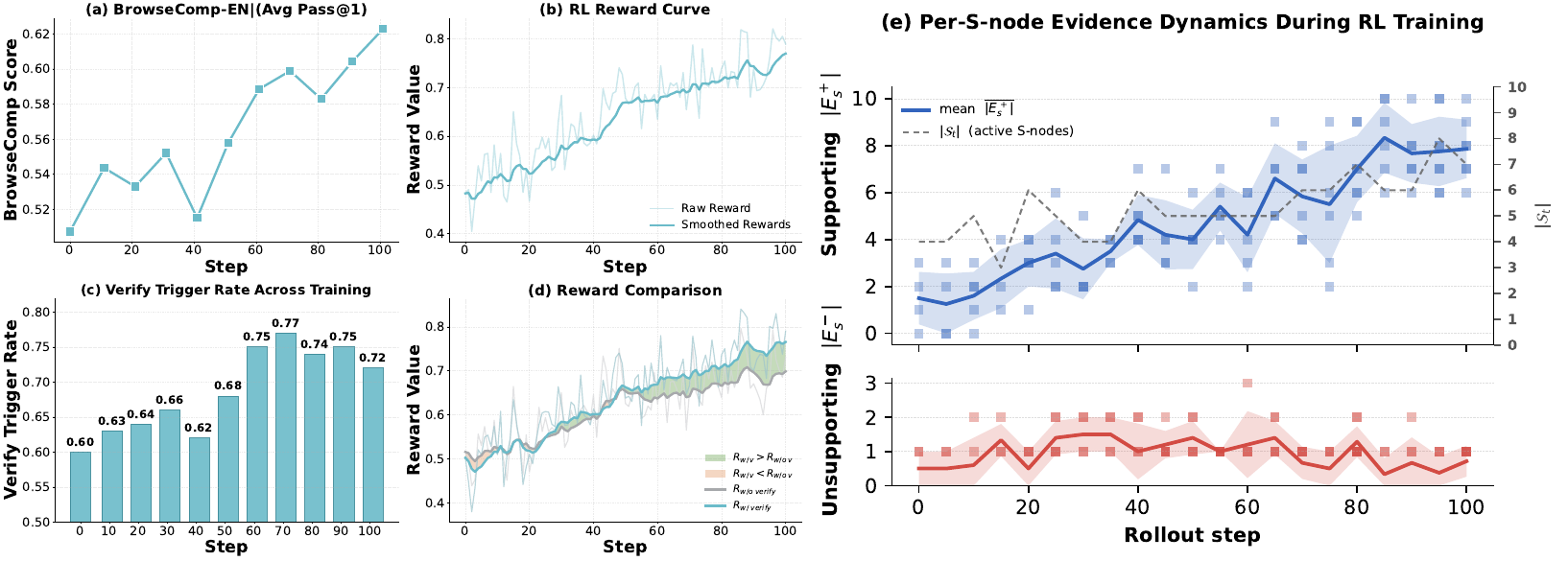}
  \caption{%
    Synthesis and verification improve jointly during GRPO training. (a) Argus-Solo accuracy on BrowseComp sample questions used for training-time monitoring only. (b)–(e) RL reward, verify trigger rate, with/without verification reward, and per-S-node evidence dynamics on the training set. $R_{\text{w/ v}}$ stays above $R_{\text{w/o v}}$ throughout training, indicating verification provides a persistent gain.
    }
  \label{fig:training_dynamics}
\end{figure}

\textbf{Training Dynamics of Verification and Synthesis}.
Figure~\ref{fig:training_dynamics} shows four trends. \emph{Reward and
accuracy track each other} (a, b): held-out BrowseComp accuracy follows
RL reward, indicating clean signal transfer. \emph{Verification gives a
persistent gain} (d): $R_{\text{w/ v}}$ stays above $R_{\text{w/o v}}$
throughout, with the gap widening over training. \emph{The Navigator
learns when to verify} (c): the verify trigger rate rises and plateaus
as synthesis nodes grow. \emph{Evidence broadens and deepens} (e):
supporting evidence per active S-node accumulates while the active set
expands, and unsupporting evidence shows a transient mid-training hump
before contracting, consistent with an exploration-to-consolidation
transition.

\subsection{Limitation and Discussion.}
\label{limitation_discussion}
Argus is designed as a heavy duty research solver rather than a low cost or low latency assistant. We acknowledge the substantial inference budget required for our approach. The cumulative Searcher token consumption per question grows from 0.4M at $K=1$ to 25.6M at $K=64$ (Figure~\ref{fig:scaling_curve}). At high $K$, the slowest Searcher in the parallel batch dominates the wall clock time, whereas the Navigator synthesis pass remains a single forward computation. We deliberately trade this high test time compute for superior accuracy and graceful scaling. Furthermore, Argus naturally inherits the recall ceiling of the Searcher when underlying web sources are absent or paywalled. The backbone transfer results in Table~\ref{tab:searcher_generalization} demonstrate that stronger Searchers lift this ceiling roughly linearly. Argus inherits standard agentic risks, including misinformation and copyright concerns; 
however, our per-claim source tracing partially mitigates misuse by ensuring strict auditability.

%% file: section/02_related_work.tex
\section{Related Work}
\label{sec:related}

\paragraph{Deep Research Agents.}
Recent deep research systems such as OpenAI Deep
Research~\cite{openai2025deepresearch}, Gemini Deep Research~\cite{google2025geminideepresearch},
WebWalker~\cite{wu2025webwalker},
WebThinker~\cite{li2026webthinker},
Webdancer~\cite{wu2026webdancer},
WebResearcher~\cite{webresearcher},
MiroThinker~\cite{mirothinker}, and Tongyi
DeepResearch~\cite{tongyi} follow a single ReAct-style
agent~\cite{react} that accumulates evidence along one sequential
trajectory. Training-based approaches such as
WebSailor~\cite{li2025websailor,websailorv2} and
WebExplorer~\cite{liu2025webexplorer} address this limitation by
training agents on high-uncertainty synthetic trajectories.
Context-management extensions such as AgentFold~\cite{ye2025agentfoldlonghorizonwebagents} and
ReSum~\cite{wu2026resumunlockinglonghorizonsearch} compress long trajectories within a single agent
to relieve context-window pressure. All of these efforts improve
what a single trajectory can reach. Argus is complementary: it
composes evidence \emph{across} multiple trajectories through an
explicit structured state, leaving the per-trajectory Searcher
untouched.

\paragraph{Parallel Agents.}
Scaling inference-time compute by running multiple parallel actors has
been pursued from two angles. The first scales chain-of-thought
reasoning through parallel sampling: self-consistency~\cite{wang2023selfconsistency},
best-of-$N$ selection~\cite{bestofn}, process reward
models~\cite{prm}, and tree-search methods~\cite{mcts}. Recent work
extends this pattern to agentic search through majority voting,
best-of-$N$ over final answers, and learned aggregation over completed
rollouts~\cite{aggagent,zeng2026pushing}. Asymmetric
verification~\cite{zeng2026pushing} in particular allocates a separate
verifier to score completed rollouts, exploiting that verification is
easier than generation. The second angle coordinates several
specialized agents toward a joint task: role-based frameworks such as
AutoGen~\cite{autogen}, CAMEL~\cite{camel}, and MetaGPT~\cite{metagpt};
LLM debate and society-of-mind approaches~\cite{llmdebate,societyofmind};
and agent-swarm architectures such as Kimi K2.5~\cite{kimik25} that
self-direct sub-agents through parallel reinforcement learning. Both
traditions consume parallel compute first and aggregate later, which
bounds the gain when $K$ actors retrieve overlapping evidence. Argus
shares the verification-as-leverage intuition of asymmetric
verification but operates in-loop, so verifier feedback shapes
which evidence gets gathered next rather than only scoring completed
trajectories. More broadly, it allocates parallel compute during search at distinct gaps in a shared evidence graph, shifting the
central operation from trajectory selection to evidence composition.

%% file: section/07_conclusion.tex
\section{Conclusion}
\label{sec:conclusion}

Parallel test-time scaling for deep research is limited not by the compute budget, but by how that compute is allocated. Sampling independent rollouts and aggregating them post hoc saturates due to overlapping evidence, and is ultimately capped by the aggregator's context window. Argus instead treats the parallel budget as a joint assembly problem. Each Searcher closes a specific gap in a shared evidence graph, which directly decouples the Navigator's context from the Searcher count. This shifts the central operation from trajectory selection to evidence composition, allowing Argus to scale to budgets where consume-then-aggregate baselines fail. Consequently, Argus reaches state-of-the-art accuracy on five of eight benchmarks and maintains log-linear scaling through $64$ parallel Searchers, compressing $25.6$M tokens of accumulated Searcher output into a $21.5$K-token graph view. We view this compositional allocation as a primary mechanism for scaling future information-seeking agents, which inherently yields fully auditable and source-traced answers.

%% file: section/A_training.tex
\section{Training Details}
\label{app:training}

\paragraph{Searcher} The Searcher shares the Navigator Qwen3.5-35B-A3B~\cite{qwen35blog} base which is a 256 expert top 8 MoE checkpoint with 35B total and 3B active parameters. It is fine tuned with supervised learning on approximately 10K trajectories synthesized via the WebSailor v2 pipeline. No Argus specific reinforcement learning is applied.

\paragraph{Navigator Base and SFT Warm Up} The Navigator is initialized from the same Qwen3.5-35B-A3B checkpoint and warm-started by SFT on graph-extraction and synthesis traces, with learning rate $1\!\times\!10^{-5}$ and batch size 64. The checkpoint with the lowest held-out loss is used to initialize RL.

\paragraph{Navigator RL Data} RL is carried out on 5298 multi hop information seeking questions. Each is annotated with a verified answer and a pre collected Searcher trajectory used as the fixed input $\mathcal{T}$. To prevent contamination we perform entity level decontamination of the training set against all eight evaluation benchmarks. Any training question whose set of named entities overlaps with that of any test question is removed prior to training. Evaluation during training is on a 200 question held out subset disjoint from all evaluation benchmarks.

\paragraph{Navigator RL Rollouts} Each training rollout is a single token sequence containing the observation stage, which builds $\mathcal{G}_{\text{pre}}$ from the
trajectory in a sliding window of 15 rounds with at most 8 windows. This matches the window range used in SFT data construction. The sequence then includes the verification stage and the synthesis stage that produces $y^{\star}_{\text{w/v}}$ over $\mathcal{G}_{\text{post}}$. The shadow synthesis $y^{\star}_{\text{w/o v}}$ over $\mathcal{G}_{\text{pre}}$ is computed only for the contrastive reward and does not enter the training sequence. Only Navigator generated tokens carry gradients. The trajectory along with verification returns and any other external input are masked from the loss. During RL we enforce a strict state machine over the DAG output format. Rollouts that violate the format are rejected before proceeding which prevents format degeneration during policy updates.

\paragraph{GRPO Hyperparameters} GRPO is run with a constant learning rate of 0.000001. The setup uses a rollout batch of 64 prompts with $N=8$ rollouts per prompt for an effective batch of 512 samples. We use an over sampling batch of 128 and a rollout temperature of 1.0. We set $\epsilon=0.2$ and $\beta=0.001$ following GRPO practice~\cite{shao2024deepseekmathpushinglimitsmathematical}. The verify bonus coefficient is $\lambda=0.5$ and the maximum response length is 135168 tokens. We train for 100 rollout steps with sample and verify timeouts of 600 seconds and 1200 seconds respectively. 

\paragraph{Compute} All training runs on 64 H200 GPUs. The 100 step GRPO run takes approximately 1.5 days of wall clock time end to end including rollout generation and policy updates.

\section{Evaluation Setup}
\label{app:evaluation}

\paragraph{Inference and Reward Constants} The per-query compute budget $B$ equals the
maximum number of Searcher dispatches, e.g., $B=64$ in the largest scaling configuration. The RL reward uses DeepSeek-V3.1-Chat as the LLM judge.

\paragraph{Evaluation Protocol} We report Pass@1 accuracy on every benchmark and follow the official evaluation protocol of each. BrowseComp and BrowseComp ZH alongside GAIA and Seal 0 and xbench DeepSearch and HLE and the two FrontierScience tracks each prescribe their own LLM as judge with a benchmark specific rubric. We adopt these without modification so our numbers are directly comparable to officially reported results. The LLM-as-judge used inside the GRPO reward is a separate training-only judge. External baseline numbers are taken from prior work as reported, except entries marked with a dagger which we reproduced ourselves under the official judge of each benchmark. The RL training set is decontaminated against all eight benchmarks at the entity level.

%% file: section/case_1.tex
\section{Case Study}
\label{showcase1}

\begin{tcolorbox}[
    enhanced, breakable,
    colback=argusbg,
    colframe=argusborder,
    boxrule=0.6pt,
    arc=4pt,
    left=8pt, right=8pt, top=6pt, bottom=6pt,
    toptitle=6pt, bottomtitle=6pt,
    titlerule=0pt,
    fonttitle=\bfseries\large,
    title={Case Study 2: Verification overturns a confidently wrong name},
    coltitle=black,
    colbacktitle=argusbg,
]

\setlength{\parindent}{0pt}
\setlength{\parskip}{4pt}

\textbf{Question.}
What is the name of the individual who received a ``special mention''
in the acknowledgments section of a thesis submitted between 2018 and
2020 to a US university established in the 1800s? The author's
then-graduate advisor was listed in the Providence Business News 40
Under Forty list. The author received the Saint Elmo Brady Award for
Outstanding Achievement in Science from the URI Black Scholar Awards
Committee. The individual contributed to a chapter of two books, both
published in 2017.

\textbf{Ground truth:} Nicholas Constant.

{\color{argusborder}\hrulefill}

\textbf{Searcher trace $H_1$.}

\begin{Verbatim}[fontsize=\small, commandchars=\\\{\}]
[thought] Five overlapping constraints. Start with the most
          distinctive: the Saint Elmo Brady Award at URI.
[search]  "URI Saint Elmo Brady Award recipient"
[obs]     ThankGod Ugochukwu (2018); Deborah Adenakan.
[search]  "URI Saint Elmo Brady Award PBN 40 Under Forty"
[obs]     Mentions Jesse Duroha alongside Gretchen Macht (PBN 2019).
[visit]   bipartisanpolicy.org/person/gretchen-macht/
[obs]     Confirms Macht received PBN 40 Under Forty in 2019.
          ... 4 rounds elided ...
[answer]  \textcolor{argusred}{Jesse Duroha.}
\end{Verbatim}

\textbf{Navigator parses $H_1$ into $\mathcal{G}_1$:}

Evidence: $E_1$ (Ugochukwu 2018), $E_2$ (Adenakan), $E_3$ (Roxbury PBN
2022), $E_4$ (Macht PBN 2019), $E_5$ (Duroha SEB).

Claim: $S_1 = $ ``Jesse Duroha is the answer.''
Status: $\sigma(S_1) = 0$ (corroborated by 2 sources, but 2018--2020
window and 2017 book chapters not touched).

Verify queries:

$v_1$: confirm Jesse Duroha satisfies all five constraints.

$v_2$: actual submission year of Duroha's thesis.

\textcolor{arguscyan}{$v_3$ (alternative-hypothesis): are there other
URI graduate students whose advisor is on PBN 40 Under Forty and who
received the Saint Elmo Brady Award?}

{\color{argusborder}\hrulefill}

\textit{[verification batch dispatched, $|\mathcal{V}|=3$; the
trace responding to $v_3$ is shown below]}

\textbf{Searcher trace $H_2$ (responding to $v_3$).}

\begin{Verbatim}[fontsize=\small, commandchars=\\\{\}]
[thought] Probe asks for OTHER URI students with the same award.
          Pivot away from Duroha; search the URI thesis archive.
[search]  "URI master thesis 2019 Saint Elmo Brady acknowledgments"
[obs]     Hit: Rassoul Diouf, "Data Analytics for Wearable IoT-Based
          Telemedicine", URI 2019.
[visit]   digitalcommons.uri.edu/.../article=2734&context=theses
[obs]     Acknowledgments paragraph: "Special mention to Nicholas
          Constant for ongoing help in the lab..." Advisor: Prof.
          Kunal Mankodiya.
          ... 3 rounds elided ...
[search]  "Kunal Mankodiya PBN 40 Under Forty"
[obs]     PBN 2017 honoree, director of URI Wearable Biosensing Lab.
[search]  "Nicholas Constant 2017 book chapter"
[obs]     Two 2017 chapters in Springer healthcare volumes.
[answer]  \textcolor{arguscyan}{Nicholas Constant.}
\end{Verbatim}

\textbf{Navigator updates $\mathcal{G}_1 \to \mathcal{G}_2$:}

New evidence: $E_6$ (Duroha thesis is 2023, contradicts window),
$E_9$ (acknowledgments paragraph, decisive),
$E_{10}, E_{15}$ (Mankodiya PBN 2017),
$E_{12}, E_{13}$ (Diouf SEB 2017),
$E_{11}, E_{14}$ (Constant 2017 chapters).

New arcs to $S_1$: $E_6, E_8, E_{16}$--$E_{19}$ all attach as
\textcolor{argusred}{contradict}, flipping $\sigma(S_1) \leftarrow -1$.

New claim: $S_3 = $ ``Nicholas Constant is the answer.''
Status: $\sigma(S_3) = +1$, supported by one evidence node per
constraint, no contradictions.

{\color{argusborder}\hrulefill}

\textbf{Proof chain over $\mathcal{G}_2$.}

The first four constraints in $q$ jointly identify a single thesis.
Within URI (a US university founded in 1892), the 2017 Saint Elmo
Brady awardee is Rassoul Diouf ($E_{12}, E_{13}$), and Diouf's
master thesis was submitted in 2019, inside the 2018--2020 window
($E_9$). The same thesis ($E_9$) names Kunal Mankodiya as the
graduate advisor, and Mankodiya is independently confirmed as the
2017 PBN 40 Under Forty honoree ($E_{10}, E_{15}$). The author and
the thesis are therefore uniquely pinned.

The fifth constraint then resolves the answer. The acknowledgments
section of Diouf's thesis, retrieved via a direct \texttt{visit}
($E_9$), gives a special mention to Nicholas Constant. Independent
sources show that Constant contributed to two 2017 book-chapter
publications in the Springer healthcare volumes
($E_{11}, E_{14}$), satisfying the final constraint.

\textbf{Final Answer}: {\textbf{Nicholas Constant}}.
\end{tcolorbox}